\documentclass[format=acmsmall, review=false, screen=true]{acmart}

\usepackage{booktabs} 

\usepackage[ruled]{algorithm2e} 
\usepackage{amsmath}
\usepackage{bbm}
\usepackage{url}
\usepackage{subfigure}

\SetAlFnt{\small}
\SetAlCapFnt{\small}
\SetAlCapNameFnt{\small}
\SetAlCapHSkip{0pt}
\IncMargin{-\parindent}

\acmJournal{TALLIP}

\setcopyright{acmlicensed}

\acmDOI{0000001.0000001}

\begin{document}
\title{$\mathbf{\mu}$-Forcing: Training Variational Recurrent Autoencoders for Text Generation}

\author{Dayiheng Liu}
\orcid{0000-0002-8755-8941}
\affiliation{
  \institution{College of Computer Science, State Key Laboratory of Hydraulics and Mountain River Engineering, Sichuan University}
  \city{Chengdu}
  \postcode{610065}
  \country{China}}
\email{losinuris@gmail.com}

\author{Yang Xue}
\authornote{Equal contribution}
\affiliation{
  \institution{College of Computer Science, Sichuan University}
  \city{Chengdu}
  \country{China}}

\author{Feng He}
\affiliation{
  \institution{College of Computer Science, Sichuan University}
  \city{Chengdu}
  \country{China}}
  
\author{Yuanyuan Chen}
\authornote{This is the corresponding author}
\affiliation{
  \institution{College of Computer Science, Sichuan University}
  \city{Chengdu}
  \country{China}}
\email{chenyuanyuan@scu.edu.cn}

\author{Jiancheng Lv}
\authornote{This is the corresponding author}
\affiliation{
  \institution{College of Computer Science, State Key Laboratory of Hydraulics and Mountain River Engineering, Sichuan University}
  \city{Chengdu}
  \country{China}}
\email{lvjiancheng@scu.edu.cn}

\begin{abstract}
It has been previously observed that training Variational Recurrent Autoencoders (VRAE) for text generation suffers from serious uninformative latent variables problem. The model would collapse into a plain language model that totally ignore the latent variables and can only generate repeating and dull samples. In this paper, we explore the reason behind this issue and propose an effective regularizer based approach to address it. The proposed method directly injects extra constraints on the posteriors of latent variables into the learning process of VRAE, which can flexibly and stably control the trade-off between the KL term and the reconstruction term, making the model learn dense and meaningful latent representations. The experimental results show that the proposed method outperforms several strong baselines and can make the model learn interpretable latent variables and generate diverse meaningful sentences. Furthermore, the proposed method can perform well without using other strategies, such as KL annealing.\footnote{Our code and data are available at \url{https://github.com/dayihengliu/Mu-Forcing-VRAE}}
\end{abstract}



\maketitle

\renewcommand\shortauthors{Liu, D. et al}

\section{Introduction}
Natural language generation has been a popular research topic over the past decades. Unsupervised learning plays an important role in this field. In unsupervised settings, the standard RNN-based models such as RNN-based language models \cite{sundermeyer2012lstm} and sequence auto-encoders \cite{dai2015semi} generate each word of a sentence conditioned on its previous generated words and hidden state. However, they do not explicitly include latent variables to capture meaningful latent features and represent the full sentence. As discussed by \cite{bowman2015generating}, these RNN-based models do not generally learn smooth and interpretable latent variables for sentence representation, which is often the main purpose of unsupervised learning. Their sentence encoding vectors cannot be used to sample novel sentences for RNN decoders.

As one kind of generative model, Variational Autoencoders (VAEs) \cite{kingma2013auto,rezende2014stochastic}, have shown great promise in image and text generation. The VAEs integrate stochastic latent variables $z$ into the auto-encoder architecture. By imposing a prior standardized normal distribution on the latent variables, the VAEs learn latent variables not as single isolated points, but as soft dense regions in latent space which makes it be able to generate plausible examples from every point in the latent space. The VAE models have been successfully used to generate plausible images \cite{yan2016attribute2image,gregor2015draw}. However, it often performs poorly on text generation. 

For text generation, autoregressive density estimators such as LSTM RNNs \cite{hochreiter1997long}, which are highly expressive, are usually employed as the decoder parts of the VAE-based models. Such models are called Variational Recurrent Autoencoders (VRAEs) \cite{fabius2014variational}. VRAEs tackle the problem of controlled generation of text. They are able to generate realistic sentence examples as if they are drawn from the input data distribution by simply feeding noise vectors through the decoder. Additionally, the latent representations obtained by applying the encoder to input examples give fine-grained control over the generation process that is harder to achieve with more conventional autoregressive models. These latent variables make it possible to control various fine-grained attributes over the generation process, such as controlling the sentiment or writing style of generated sentences.

Nevertheless, VRAEs face some optimization challenges. As argued in \cite{bowman2015generating,chen2016variational,zhao2017infovae,alemi2018fixing}, the core difficulty of training VRAEs is that the models would suffer from serious uninformative latent variables (also called KL vanishing) issue: the VRAEs tend to totally ignore the latent variables and only use the decoder part to model the data. In practice, the VRAEs would collapse into plain language models and can only generate repeating and dull samples.

To mitigate this uninformative latent variables problem, \cite{bowman2015generating} propose a trick called KL cost annealing. However, the training of VRAEs still be prone to collapse on large corpus with this trick. As pointed by \cite{yeung2017tackling}, this hand-tuned method make the process of training VRAEs still difficult and are not very efficient.

We propose a regularizer based approach called $\mu$-Forcing to address the uninformative latent variables problem. An additional regularizer is added to the objective function of original VRAE which prevents the VRAE collapses into a trivial solution and guides the VRAE to explore its model capacity to learn a better latent representation. This method stems from the following intuition: when the model collapses into a trivial solution, the approximation posterior distribution $q(z | x)$ of every data point $x$, which is usually assumed to be $\mathcal{N}(\mu, \sigma^2)$, degrades to the same $\mathcal{N}(0, 1)$. However, for reasonable latent representations, different data points $x$ should have different latent representations $q(z | x)$. The proposed method introduces a mild constraint on the $\mu$ of $q(z | x)$ to force the model to find a non-trivial solution where the learned latent variables $z$ contain useful information. 

Specifically, the contributions of this paper can be summarized as follows:
\begin{itemize}
\item We propose an effective method to address the uninformative latent variables problem for VRAEs. This method can flexibly and stably control the trade-off between the KL term and the reconstruction term, making the model learn dense and meaningful latent representations. Furthermore, our proposed method can perform well without using other strategies, such as KL annealing.
\item For sentence generation, the experiments indicate that our proposed method outperforms several strong baselines. We show that our method can generate diverse meaningful sentences and learn interpretable latent variables.
\end{itemize}

\section{Related Work}
When applied to text generation or complex datasets such as ImageNet \cite{deng2009imagenet}, VAEs suffer from two major problems: blurry samples and uninformative latent variables. Lots of approaches have been proposed to address these issues. Our work falls into this category, but focuses on text generation where the second issue dominates.

Recently, much work has been done to come up with more powerful posterior distributions. \cite{rezende2015variational} learn highly non-Gaussian posterior densities by transforming simple densities into complex ones with sequences of invertible transformations. \cite{makhzani2015adversarial} introduce generative adversarial networks (GAN) \cite{Goodfellow2014Generative} to variational inference, which can match the posterior distribution with an arbitrary prior distribution. These methods have shown to be effective in improving variational inference and solving the blurry sample issue partially. However, as \cite{sonderby2016ladder,bowman2015generating} observed, these approaches seem to do little on the uninformative latent variables problem.

The uninformative latent variables issue is formally studied in \cite{chen2016variational}, which casts the problem of optimizing VAE into designing an efficient coding scheme. Their work shed light on the reason of the uninformative latent variables problem from the perspective of coding theory, but without proposing any principle method to address it. \cite{zhao2017infovae} further formally study these problems of VAE, and propose a family of VAE based models. They demonstrate that all of these models maximize the mutual information between input and latent variables and achieve better performance on image generation.

As for sentence generation, recent attempts that use autoregressive conditional likelihood in VAEs suffer from seriously uninformative latent variables issue. Existing solutions to this problem can be divided into two categories: model-based and regularizer-based methods. For model-based methods, \cite{semeniuta2017hybrid} proposes a novel hybrid convolutional-recurrent model (hybrid-VAE) with an additional auxiliary reconstruction term to address the uninformative latent variables issue for text generation. This architecture is attractive for its computational efficiency but less flexible. \cite{yang2017improved} extends the hybrid-VAE by introducing dilated convolutions to improve the variational model for text generation. \cite{goyal2017z} proposes a stochastic recurrent model in which each step in the sequence is associated with a latent variable. To ease the training, they add an auxiliary cost to force each latent variable to reconstruct the state of the backward recurrent network. In order to solve the KL vanishing and inconsistent training objective for dialogue generation, \cite{shen2018improving} firstly learns to autoencoder discrete texts into continuous embeddings which are sampled by transforming Gaussian noise and are trained with a separate model. Then the model learns to generalize latent representations by reconsturcting the encoded embedding.

For the regularizer-based method, \cite{bowman2015generating} uses the KL cost annealing method to enforce the VRAEs to learn to encode as much information in latent variables as it can in the early stage of training. In addition, they weaken the decoder by randomly removing some conditional information (the ground-truth previous word) during training to force the model to rely on the latent variables. In practice, these tricks are not always effective. Another popular strategy is free bits \cite{kingma2016improved}. This method reserves some space of KL divergence for every dimension of latent variables. Similarly, \cite{yang2017improved} reserves space for the total KL divergence instead of for for every dimension. In order to solve the uninformative latent variables problem, \cite{zhao2017learning} forces the latent variables to predict the bag-of-words vector of the reconstruction sentence. Nevertheless, this method needs to incorporate another neural network to predict the bag-of-words vector, which will significantly increase the number of parameters of the model. More recently, \cite{alemi2018fixing} presents a theoretical framework for understanding representation learning using latent variable models in terms of the rate-distortion tradeoff, and confirms that the VAE models with expressive decoders can ignore the latent code. They propose a simple solution that reduces the KL penalty term $\lambda$ to $\lambda < 1$ to this problem. However, their experiments are based on image generation, and we experimentally find that when applying this method to VRAE model for text generation, we need to carefully adjust the value of $\lambda$ and employ the KL cost annealing trick to avoid the KL vanishing problem.

Compared with the model-based methods, the regularizer-based methods are more flexible and scalable. Our approach is also a regularizer-based method which directly injects constraint on the posterior of latent variables. The experimental results demonstrate that the proposed method outperforms several regularizer-based baselines. In addition, our experiments show that the proposed method can be also applied to other model-based methods and improve the performance, such as hybrid-VAE \cite{semeniuta2017hybrid}. 

\section{Variational Recurrent Autoencoders}
VAE framework is a neural network based method for training generative latent variable models which integrates stochastic latent variables $z$ into the auto-encoder architecture. Let $x$ be an observed variable. Given a set of observed data points $\mathcal{X} = \{x^{(1)}, ..., x^{(N)}\}$, the goal is to estimate the parameters $\theta$ that maximize the marginal log-likelihood:
\begin{equation}
    \log p_{\theta}(\mathcal{X}) = \sum_{n=1}^N \log \int_{z}  p(z) p_{\theta}(x^{(n)} | z) \mathrm{d}z.  \notag
\end{equation}
In general, we assume that the prior distribution $p(z)$ is normal distribution $\mathcal{N}(z; 0,I)$. Due to the presence of integral in the marginal log-likelihood, it is intractable to directly compute or differentiate the marginal log-likelihood. A common solution is to optimize the evidence lower bound (ELBO) on the marginal log-likelihood by introducing an approximate posterior distribution $q_{\phi}(z|x)$:
\begin{align}
	\log p_{\theta}(x) \geq  & \mathbb{E}_{q_{\phi} (z|x)} \Big[\log p_{\theta} (x|z) \Big]  \notag\\
    & - \mathcal{D}_{KL} \big( q_{\phi} (z|x)\parallel p(z) \big).
\end{align}
Here two neural networks with parameter $\phi$ and $\theta$ are respectively employed for modeling the posterior distribution $q_{\phi}(z|x)$ and the conditional distribution $p_{\theta}(x|z)$. In general, we assume that $q_\phi(z|x)$ is multivariate diagonal Gaussian distribution:
\begin{equation}
q_{\phi}(z | x)  = \mathcal{N} \big(z; \mu_\phi(x), \Sigma_\phi(x) \big).  \notag
\end{equation}
where $\mu_\phi$ and $\Sigma_\phi$ are implemented via neural networks with parameters $\phi$. However, sampling $z$ from $q_{\phi}(z | x)$ is a non-continuous operation and has no gradient. The solution is the reparameterization trick which first samples $\epsilon \sim \mathcal{N}(0, I)$, and then computes $z = \mu_\phi(x) + \Sigma_\phi^{\frac{1}{2}}(x) * \epsilon$. Through this trick, the VAE can be trained by stochastic gradient descent with minimizing the objective function:
    \begin{align}
      \mathcal{L}_{vae} (\mathcal{X}; \phi, \theta) &=  \mathcal{L}_{recon} + \mathcal{L}_{KL}    \notag \\
        &=  - \sum_{n=1}^N \mathbb{E}_{q_{\phi} (z|x^{(n)})} \Big[\log p_{\theta} (x^{(n)}|z) \Big]\notag \\
	&+  \sum_{n=1}^N \mathcal{D}_{KL} \big( q_{\phi} (z|x^{(n)}) \parallel p(z) \big). \notag
    \end{align}
From this equation, the VAE can be interpreted as a regularized auto-encoder that the $q_\phi(z|x)$ is the encoder while $p_\theta(x|z)$ is the decoder. This perspective provides an intuitive explanation of why VAE works. The reconstruction term $\mathcal{L}_{recon}$ makes VAE learn to reconstruct the input, meanwhile the KL term $\mathcal{L}_{KL}$ encourages $q_\phi(z|x)$ to match the prior $p(z)$. Finally, the VAE can generate plausible samples from $p(z)$.
\begin{figure}[t]
   \centering
   \includegraphics[width=4.5in]{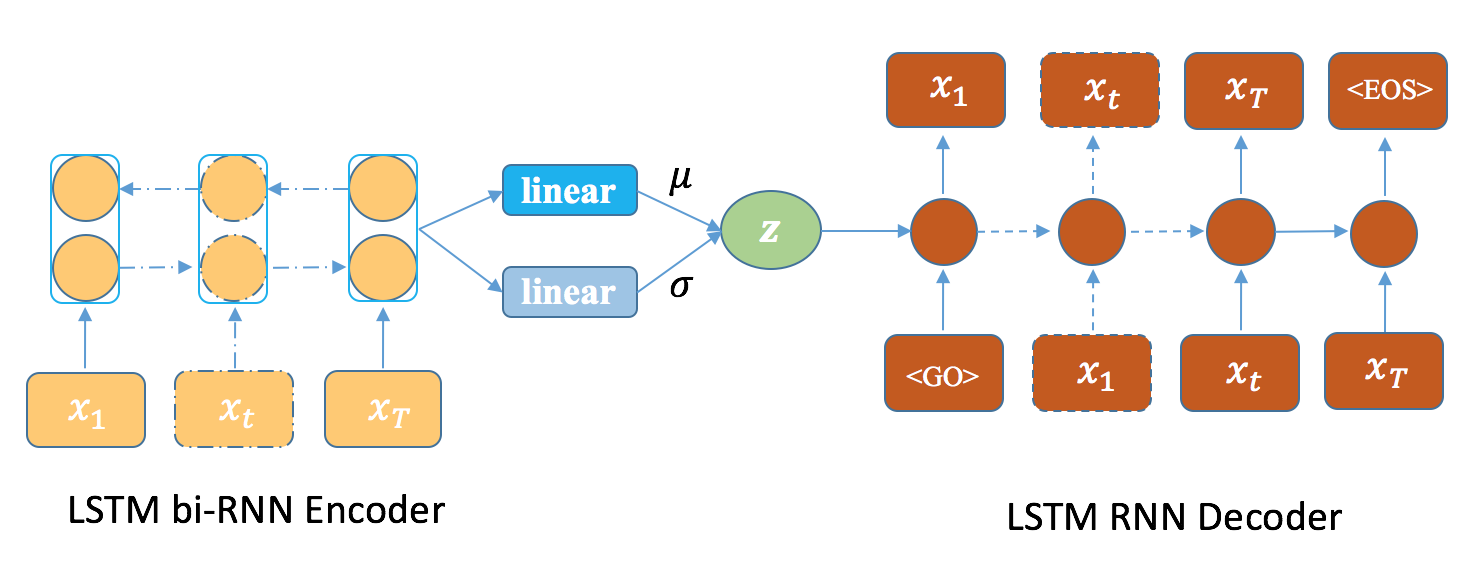}
   \caption{The architecture of VRAE that we used in this paper.}
   \label{fig:vfigure1}
\end{figure}
VRAE shares the same architecture of the VAE, but autoregressive model is employed for the posterior distribution $p_{\theta}(x|z) $ (see Figure \ref{fig:vfigure1}). Autoregressive models, such as RNNs, omit the independent assumption and predict the data with history information, which can fit arbitrary distribution in theory.

\section{$\mu$-Forcing Approach}
We firstly analyze the relation between the reconstruction term and the KL term in $\mathcal{L}_{vae}$ from the perspective of mutual information. Let us rewrite the mutual information of $x$ and $z$:
\begin{align}
    I(x, z) &= \mathbb{E}_{p(x,z)}\left[\log p\left ( x,z \right ) - \log p\left ( x \right )p\left ( z \right ) \right]  \notag\\
    &= \mathbb{E}_{p(x)}\left[ \mathcal{D}_{KL} \left[ p(z|x) \parallel p(z) \right ] \right]. \notag
\end{align}
We can see that the KL term estimates the mutual information $I(x, z)$ by empirical data distribution $p(x)$ and approximate posterior distribution $q_{\phi} (z|x)$. From this perspective, the KL term indicates how much information the VAE stores in the latent variables and minimizing it is going to penalize the $I(x, z)$. While minimizing the reconstruction term amounts to maximizing a lower bound on the mutual information $I(x, z)$, as discussed by \cite{vincent2010stacked}. Therefore, there is an adversarial relationship between these two terms in the process of optimization. Virtually the modeling power of VAE completely benefits from this competition. It makes VAE learn a dense and meaningful latent representation. However, the architecture of VRAE makes it prone to break the balance of the competition and causes the uninformative variables issue to be more serious.

As Figure \ref{fig:vfigure1} shows, the reconstruction information of VRAE comes from two parts: the encoder and the ground truth inputs of the decoder. At the early stage of training, the encoder is poorly trained. It makes the decoder tend to depend on the ground truth inputs for reconstruction, which in turn causes the encoder cannot gain sufficient reconstruction error signals against the KL term during optimizing. As a result, the tension between the reconstruction term and KL term is vanishing and the optimization process is divided into two unrelated parts. For the encoder, the KL term is dominated. For the decoder,
the latent variable is ignored.

The KL term $\mathcal{L}_{KL}$ is the KL-divergence between two multivariate Gaussian distributions which can be computed in closed form:
\begin{align}
\mathcal{L}_{KL}= \frac{1}{2}({\rm tr}(\Sigma_\phi) + \mu_\phi^\mathsf{T}\mu_\phi + \log \det(\Sigma_\phi) - K), \notag
\end{align}
where $K$ is the dimensionality of the distribution. This equation has an unique global minima $0$ at $\mu_\phi=0$ and $\Sigma_\phi=I$. Directly optimizing it causes the $\mu_\phi$ of all data points to collapse to $0$ and the $\Sigma_\phi$ to close to $I$. This is what indeed happen in practice that the latent representation $q(z|x)$ of every data point $x$ degrades to the same $\mathcal{N}(0, I)$ and the model encodes meaningless information into the latent variables. For the decoder part, when $I(x, z)=0$, thus $p(x|z) = p(x)$ and the reconstruction term is equivalent to the negative log-likelihood. As a whole, the $\mathcal{L}_{KL}=0$ is a trivial solution which learns meaningless latent variables.

As discussed above, the log-likelihood may not guide the model towards meaningful latent representations. A straightforward approach to address this issue is to inject extra constraints on the posteriors of latent variables into the learning process of VRAE. For reasonable latent representations, different data points $x$ should have different latent representations. Inspired by this intuition, we introduce an additional constraint on the $\mu_\phi$ of $q_\phi(z | x)$ that forces the VRAE to exploit its modeling power and to learn discriminated latent representations. 

Specifically, we propose a margin-based additional cost computed from a batch of data points. For simplicity, we use $\mu$ to denote the vector $\mu_\phi$, the additional cost is as follows:
\begin{equation}
    \mathcal{L}_{\mu} = max  \bigg \{0, \beta -  \frac{1}{2N}\sum_{n=1}^{N}(\mu^{(n)} - \bar \mu)^\mathsf{T} (\mu^{(n)} - \bar \mu) \bigg \} .\notag
\end{equation}
Here $\beta$ is a margin, $N$ is the batch size\footnote{In theory, the batch size should be as large as possible. In practice, we found that when the batch size was 64, 128 and 256, the model performed well.}, $n$ denotes the $n$-th sample of a batch, $\mu^{(n)}$ denotes the $\mu$ vector of the $n$-th sample, and $\bar \mu$ is the mean of the $\mu$ vectors in this batch. This cost forces the sample variance of $\mu$ to be controlled on the level of $\beta$ which maintains the mutual information of $x$ and $z$. Intuitively, the proposed $\mathcal{L}_{\mu}$ term can prevent the KL term from closing the connection between the encoder and the decoder. 

Theoretically, we analyze the effect of $\mathcal{L}_\mu$. Given a batch data, when the variance of $\mu$ is less than the threshold value $\beta$ (the $\mu$ vectors of each data are very close to each other), the $\nabla_{\mu} \mathcal{L}_{\mu} \approx  -\mu + \bar \mu$, and the $\nabla_{\mu} \mathcal{L}_{KL} = \mu$. After the introduction of $\mathcal{L}_\mu$, $\nabla_{\mu} (\mathcal{L}_{\mu}  + \mathcal{L}_{KL}) \approx \bar \mu$. We can see that for the entire dataset, the mean of vector $\mu$ will still tend to zero under the influence of the new cost function $\mathcal{L}_{\mu}$, which guarantees the properties of VAE. Furthermore, for each individual data $x$, its $\mu$ will no longer independently converge to 0. The gradient $\nabla_{\mu} (\mathcal{L}_{\mu}  + \mathcal{L}_{KL}) $ will adaptively change dynamically. As $ \bar \mu$ gets closer to zero vector, the gradient $\nabla_{\mu} (\mathcal{L}_{\mu}  + \mathcal{L}_{KL}) $ becomes smaller and smaller. This makes each $\mu$ no longer close to each other and overlaps to the same point, ensuring the diversity of $\mu$. Therefore, $\mu$ can provide meaningful information for the decoder. The final cost function is:

\begin{equation}
     \mathcal{L} (\mathcal{X}; \phi, \theta) =  \mathcal{L}_{recon} + \mathcal{L}_{KL} + \mathcal{L}_{\mu}. \notag
\end{equation}
As discussed above, this new cost function $\mathcal{L}$ can control the trade-off between KL term and reconstruction term that prevents the model collapsing to the trivial solution. It puts no limits on the expressive power of the autoregressive model, and guide the model to learn more expressive latent variables.

\begin{figure*}[t]
  \centering
  \subfigure[KL loss on COSR]{
  \begin{minipage}{4.5cm}
  \centering
    \includegraphics[width=4.5cm]{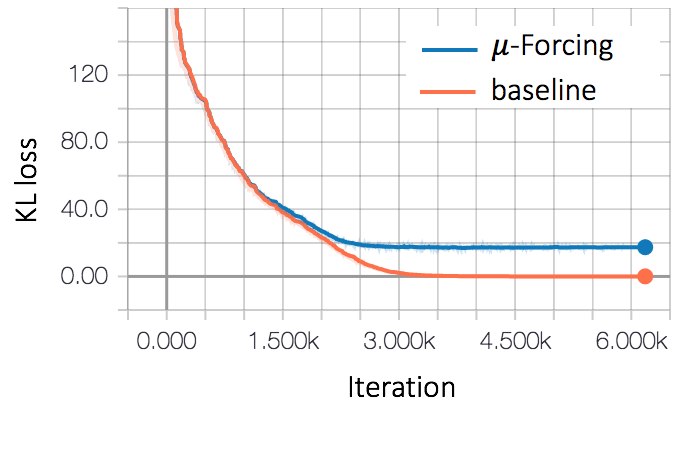}
  \end{minipage}
  \label{show1:1}
  }
  \subfigure[Reconstruction loss on COSR]{
  \begin{minipage}{4.5cm}
  \centering
    \includegraphics[width=4.5cm]{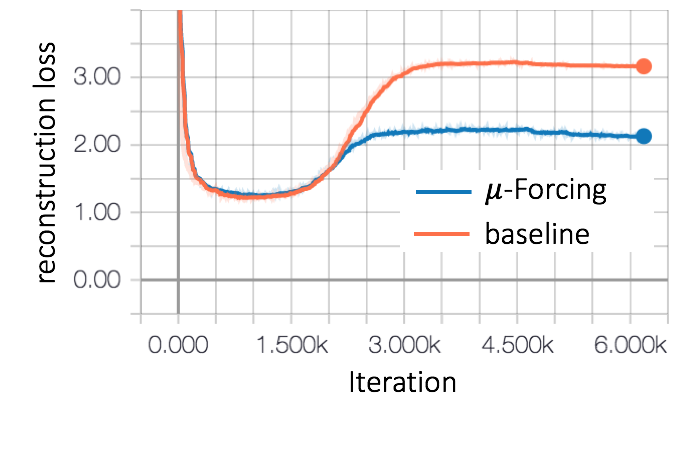}
  \end{minipage}
  \label{show1:2}
  }
 \subfigure[The distribution of $\mu$ values of baseline on COSR]{
  \begin{minipage}{4.5cm}
  \centering
    \includegraphics[width=4.5cm]{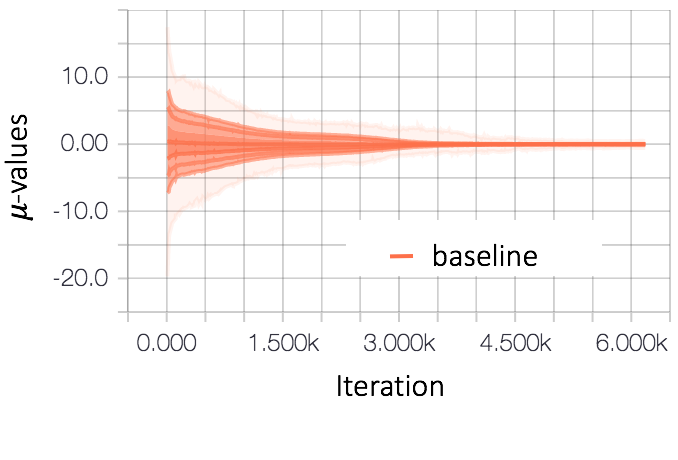}
  \end{minipage}
  \label{show1:3}
  }
 \subfigure[The distribution of $\mu$ values of $\mu$-Forcing on COSR]{
  \begin{minipage}{4.5cm}
  \centering
    \includegraphics[width=4.5cm]{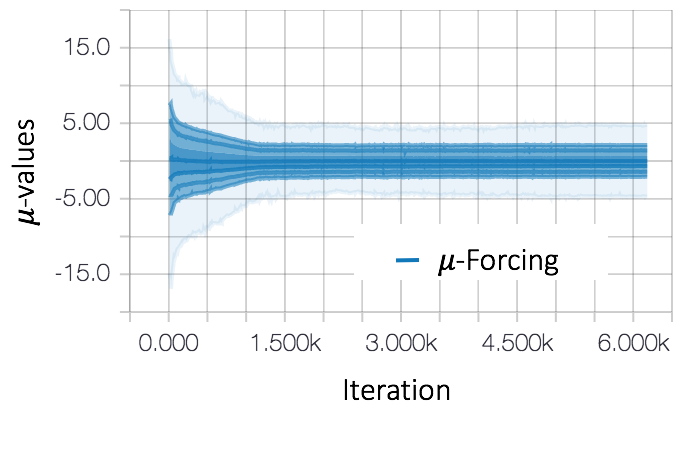}
  \end{minipage}
  \label{show1:4}
  }
  \subfigure[KL loss on APRC]{
  \begin{minipage}{4.5cm}
  \centering
    \includegraphics[width=4.5cm]{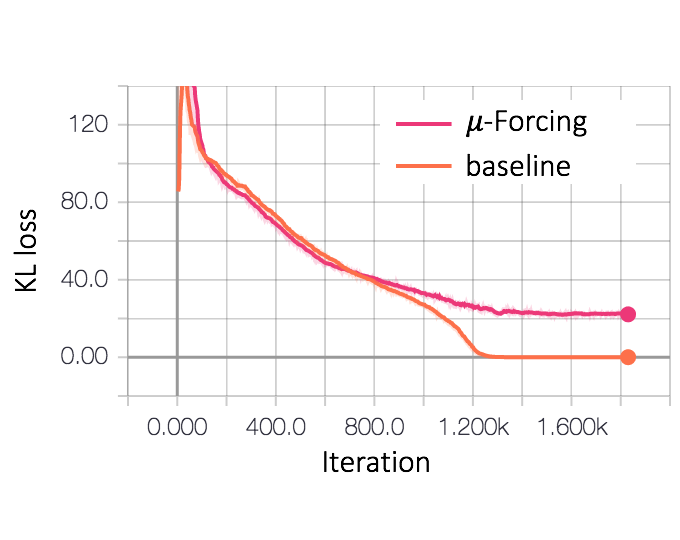}
  \end{minipage}
  \label{show2:1}
  }
  \subfigure[Reconstruction loss on APRC]{
  \begin{minipage}{4.5cm}
  \centering
    \includegraphics[width=4.5cm]{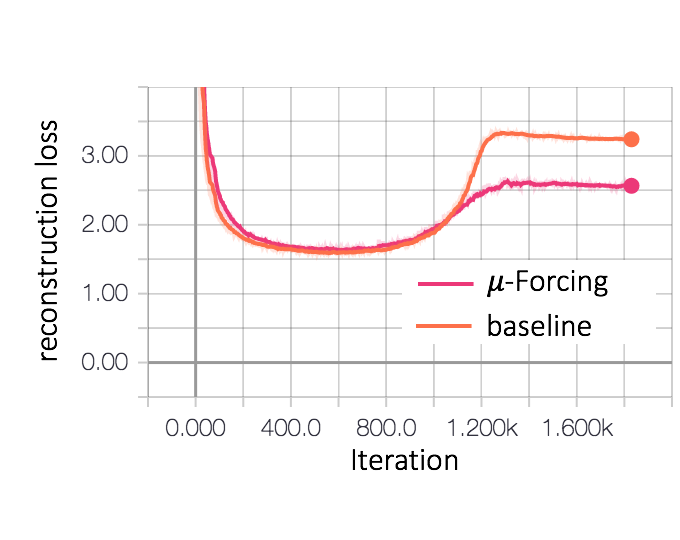}
  \end{minipage}
  \label{show2:2}
  }
 \subfigure[The distribution of $\mu$ values of baseline on APRC]{
  \begin{minipage}{4.5cm}
  \centering
    \includegraphics[width=4.5cm]{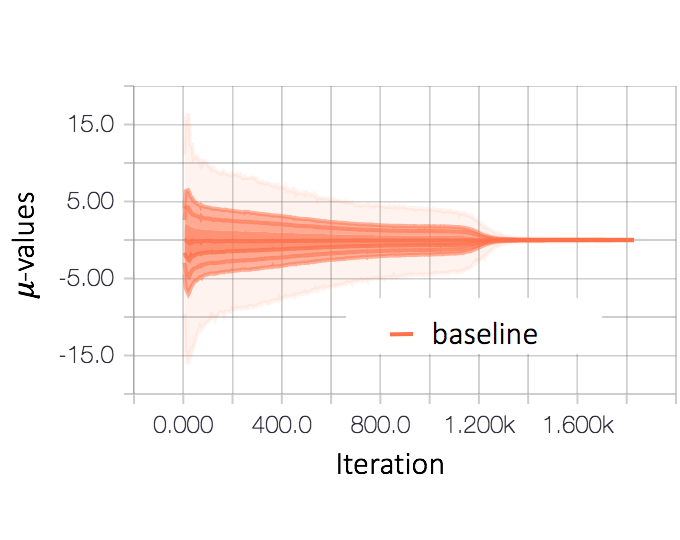}
  \end{minipage}
  \label{show2:3}
  }
 \subfigure[The distribution of $\mu$ values of $\mu$-Forcing on APRC]{
  \begin{minipage}{4.5cm}
  \centering
    \includegraphics[width=4.5cm]{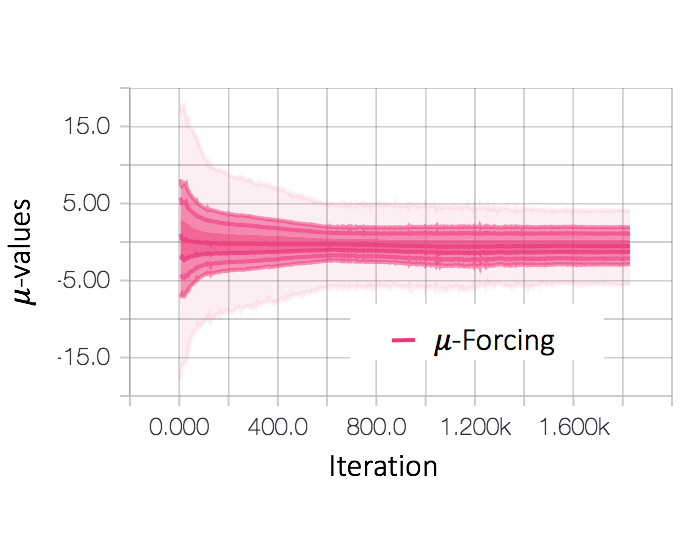}
  \end{minipage}
  \label{show2:4}
  }
 \caption{The subfigure \ref{show1:1}-\ref{show1:4} presents the results on COSR test set while the subfigure \ref{show2:1}-\ref{show2:4} presents the results on APRC test set. The subfigure \ref{show1:1} and \ref{show2:1} shows the KL loss of $\mu$-Forcing and baseline on the test set. The subfigure \ref{show1:2} and \ref{show2:2} shows the reconstruction loss (NLL) of $\mu$-Forcing and baseline on the test set. The distribution of $\mu$ values of baseline on the test set are shown in the subfigure \ref{show1:3} and \ref{show2:3}. While the distribution of $\mu$ values of $\mu$-Forcing on the test set are shown in subfigure \ref{show1:4} and \ref{show2:4} [best viewed in color].}
    \label{show1}
 \end{figure*}
\section{Experiments}
The experiments revolve around the following questions: \textbf{Q1:} Whether the proposed method can effectively solve the uninformative latent variables problem of VRAE? \textbf{Q2:} How does the parameter $\beta$ affect the model? Can this proposed method work well without using other strategy, such as KL annealing? \textbf{Q3:} How does the proposed method compare with existing state-of-the-art ones? \textbf{Q4:} In addition to sentence generation, the VRAE is mainly used to learn smooth and interpretable latent variables for sentence representation. Can the proposed method help the model learn interpretable latent variables?

\subsection{Datasets}
In the experiments, we used two medium-sized corpus. The first one is the Amazon Product Reviews Corpus (APRC) \cite{dong2017learning} which is built upon Amazon product data \cite{mcauley2015inferring}. This dataset contains 937,033 reviews and every review is paired with attributes. Here we ignored these attributes and only used the review part to train the model. We set the vocabulary size to 10K. The second dataset is Chinese Online-Shopping Reviews Corpus (COSR). We manually crawled Chinese online-shopping reviews from the internet. After cleaning the data and word segmentation, the reviews whose lengths are greater than 40 words are filtered. We finally obtained 584,475 reviews including positive, negative and neutral reviews.\footnote{In order to make our results easy to reproduce, we will release all the datasets and codes upon acceptance of the paper.} After processing low-frequency words, the vocabulary size is 9191. These two datasets are both randomly split into train/valid/test sets following these ratios respectively: 85\%, 5\%, 10\%.

\subsection{Comparison with VRAE (Q1)}
\textbf{Setup}. To answer the first question, we tested the proposed method on COSR and APRC datasets respectively. The VRAE model trained with KL annealing strategy is taken as the baseline \cite{bowman2015generating} for comparison. We compared the proposed method and baseline with the same architecture, and the only difference is the proposed method was trained with an additional cost $\mathcal{L}_{\mu}$. We used a bidirectional LSTM with 1024 hidden units for the encoder and a single layered LSTM with 1024 hidden units for the decoder. The dimension of word embeddings was set to 512 while the latent variables was set to 16. The batch size, threshold of element-wise gradient clipping and initial learning rate of Adam optimizer \cite{kingma2014adam} were set to 64, 5.0 and 0.001. We also made use of Layer Normalization \cite{ba2016layer} to make the training more easier. The hyperparameter $\beta$ of the proposed method was set to 2.

\noindent \textbf{Results}. We report the training curve of KL loss $\mathcal{L}_{KL}$, reconstruction loss $\mathcal{L}_{recon}$\footnote{In the figure, we plot the negative loglikelihood (NLL) results, which averaged the reconstruction loss by length.}, and the value distribution of vector $\mu$ on the test set of two datasets. The results are presented in Figure \ref{show1}. The first two lines present the results on COSR test set while the last two lines present the results on APRC test set. 

From the Figure \ref{show1:1} and \ref{show2:1}, we can see that the proposed method helps the model to achieve non-zero KL cost. However, even trained with KL annealing, the KL cost of the baseline still close to zero on both test sets. As the KL term annealed, we can find that the reconstruction errors on the two test sets of baseline become higher and higher from Figure \ref{show1:2} and \ref{show2:2}. Finally, the reconstruction errors of baseline are much higher than the proposed method. Furthermore, we counted the distribution of $\mu$ values on test sets. From the Figure \ref{show1:3}, \ref{show1:4}, \ref{show2:3} and \ref{show2:4}, it can be seen that most of $\mu$ values of baseline collapse to zero while the proposed method doesn't.

For the baseline, the results show that even trained with KL annealing, the VRAE still suffers from the uninformative latent variables problem on both datasets, result in high reconstruction errors and KL cost close to zero on test sets. The $\mu$ values of every data point $x$ of the baseline degrade to zero which cannot provide meaningful information. To make sure that this effect is not caused by optimization difficulties or the configuration of the model, we also searched the different hyperparameters but got the same results on both datasets. 

\begin{figure*}[h] 
   \centering
   \includegraphics[width=5.0in]{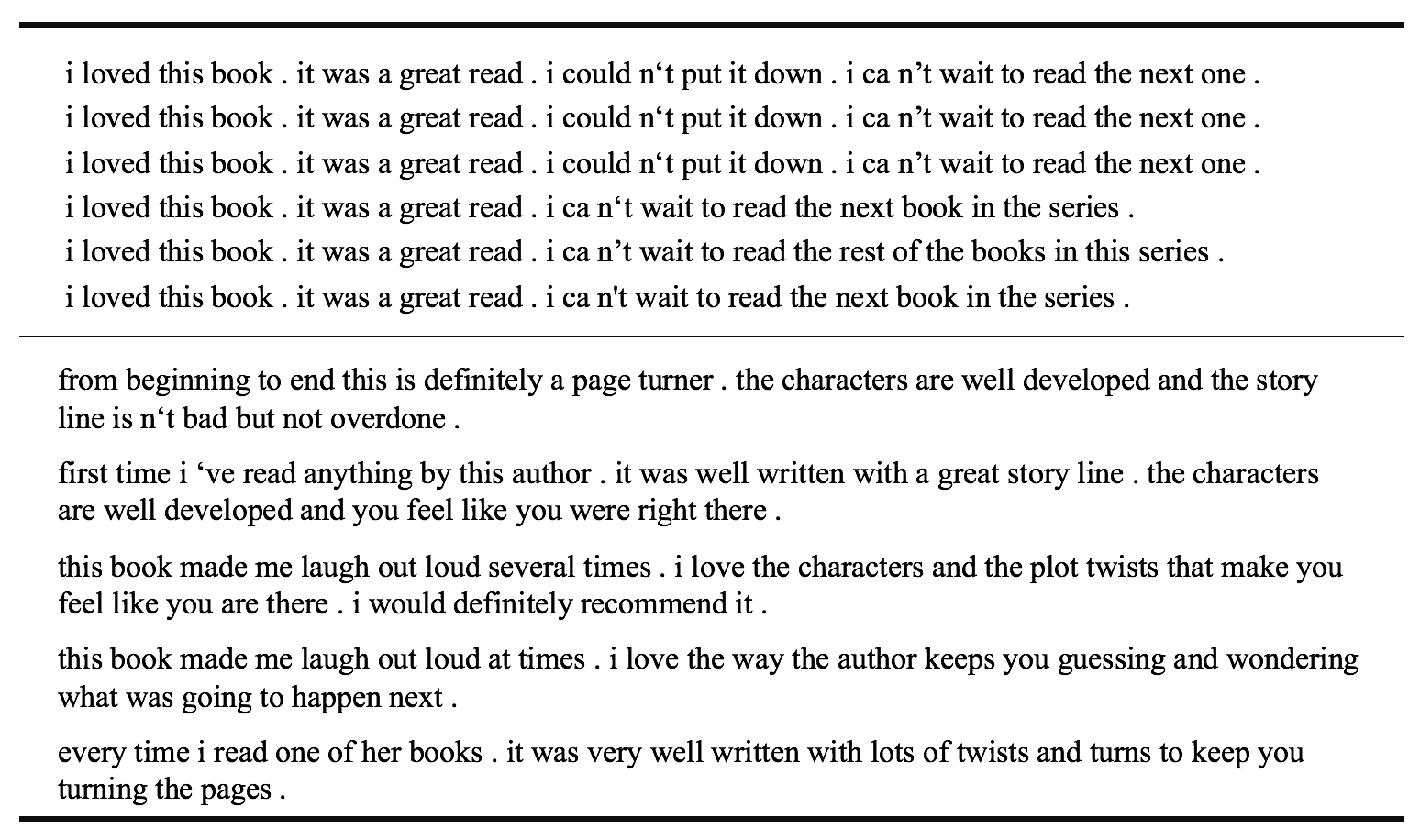}
   \caption{Examples of randomly generated by the baseline model (top section) and the proposed model (bottom), both models were trained on the APRC dataset.}
   \label{fig:afigure1}
\end{figure*}

\begin{figure*}[h] 
   \centering
   \includegraphics[width=5.0in]{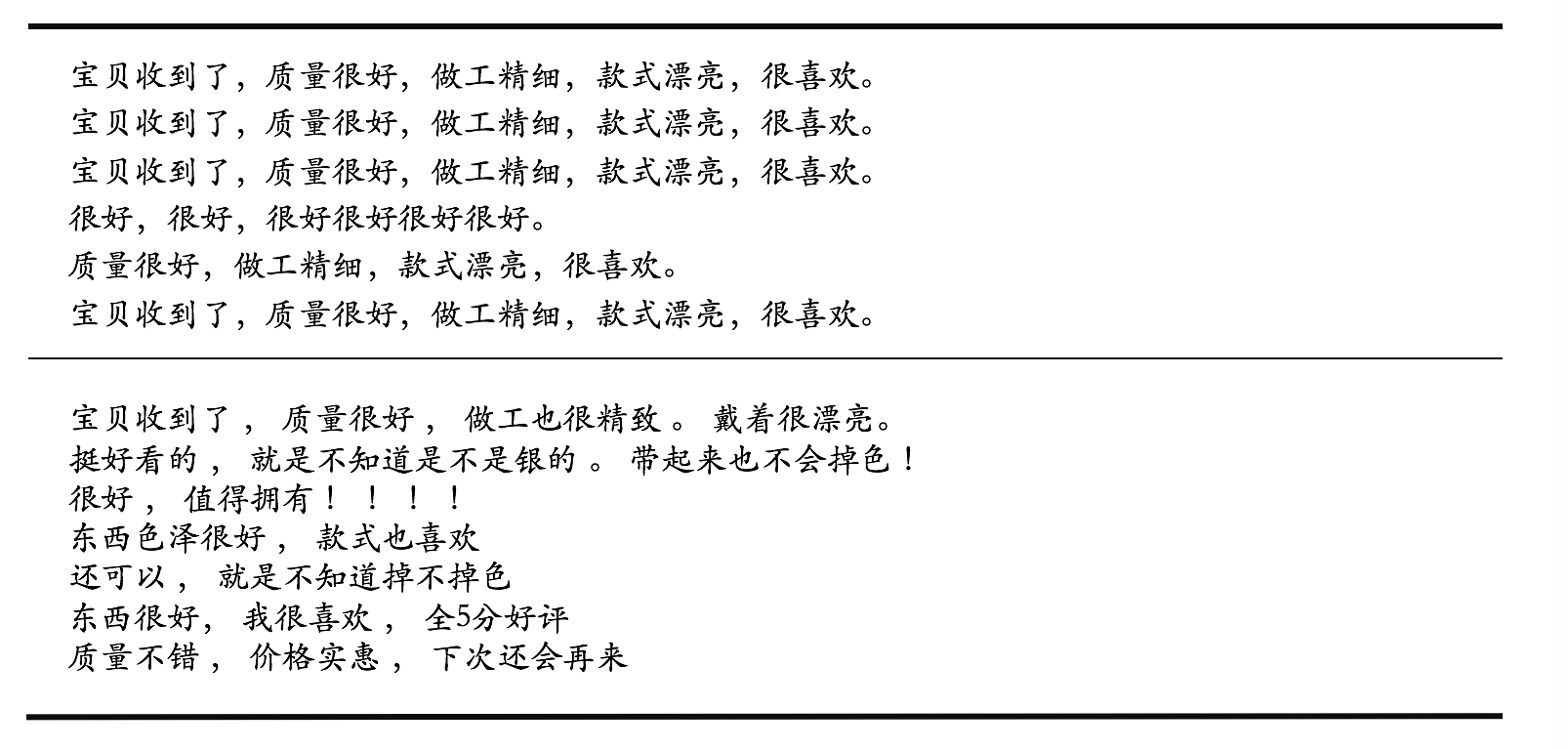}
   \caption{Examples of randomly generated by the baseline model (top section) and the proposed model (bottom), both models were trained on the COSR dataset.}
   \label{fig:afigure2}
\end{figure*}

For the proposed method, it helps the model to achieve non-zero KL term and results in lower reconstruction errors. This shows that the decoder of the proposed method utilizes the reconstruction information not only from the ground truth inputs, but also the latent variables. Otherwise, even though the proposed method holds the KL-term, the reconstruction error should be consistent with the baseline. In addition, the $\mu$ values of the proposed method are zero-centered and diverse that is what we want. It demonstrates that the proposed method can successfully solve the uninformative latent variables problem. By sampling $z$ from $\mathcal{N}(0,I)$, the baseline only generated repeating sentences while the proposed method can generate diverse meaningful sentences. We show some generated examples in the Figure \ref{fig:afigure1} and \ref{fig:afigure2}. 

It is worth noting that \cite{bowman2015generating} propose the word dropout trick to train the VRAE. We also used the word dropout trick in our experiment, but we found that it did not mitigate the KL vanishing issue. The same phenomenon has been mentioned in \cite{yang2017improved} and \cite{kim2018semi}. In addition, as discussed by \cite{semeniuta2017hybrid}, the word dropout tends to slow down convergence. It is not a stable and always effective trick. In our experiments, we also follow \cite{alemi2018fixing} to set the weight of KL cost $\lambda$ (KL penalty term) of the baseline model to less than one, for example, 0.1 or 0.3. However, with this trick, most of the time we still encounter the KL vanishing problem. We can mitigate it only by carefully adjusting the parameter values of $\lambda$ and KL annealing. We found that neither of them can guarantee a stable and effective solution to uninformative latent variables issue, but our method can easily solve the uninformative latent variables problem.

\subsection{The Effect of $\beta$ (Q2)}
\begin{figure}[t]
  \centering
  \subfigure[KL loss with various $\beta$ on COSR]{
  \begin{minipage}{4.5cm}
  \centering
    \includegraphics[width=4.5cm]{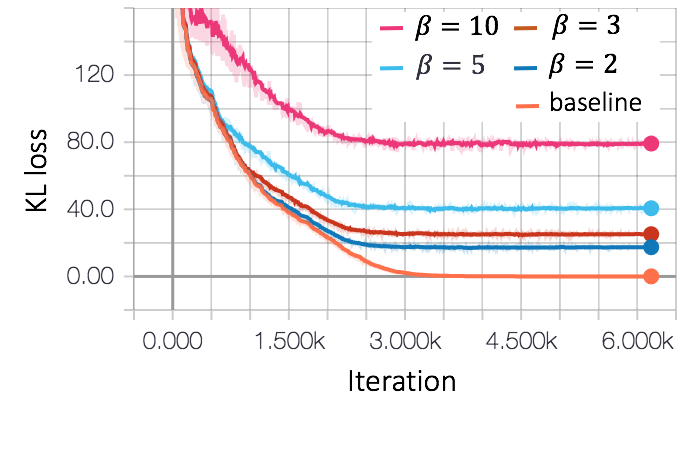}
  \end{minipage}
  }
  \subfigure[Reconstruction loss with various $\beta$ on COSR]{
  \begin{minipage}{4.5cm}
  \centering
    \includegraphics[width=4.5cm]{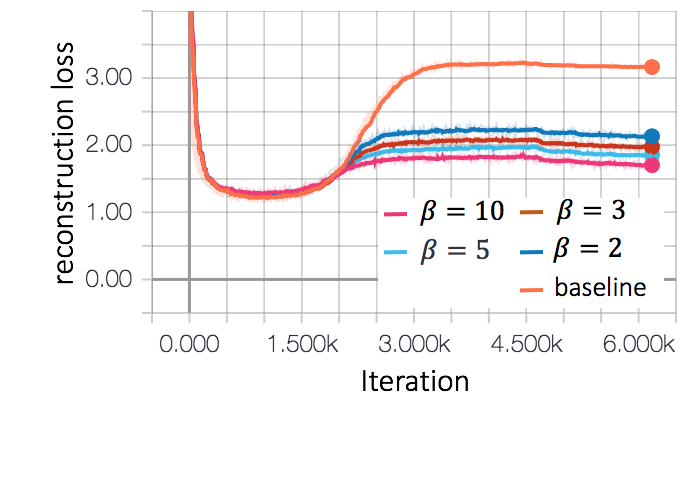}
  \end{minipage}
  }
  \caption{The KL loss and reconstruction loss of $\mu$-Forcing trained with various $\beta$ on COSR [best viewed in color].}
   \label{show}
 \end{figure}
 \begin{figure}[t]
  \centering
  \subfigure[KL loss without KL annealing on COSR]{
  \begin{minipage}{4.5cm}
  \centering
    \includegraphics[width=4.5cm]{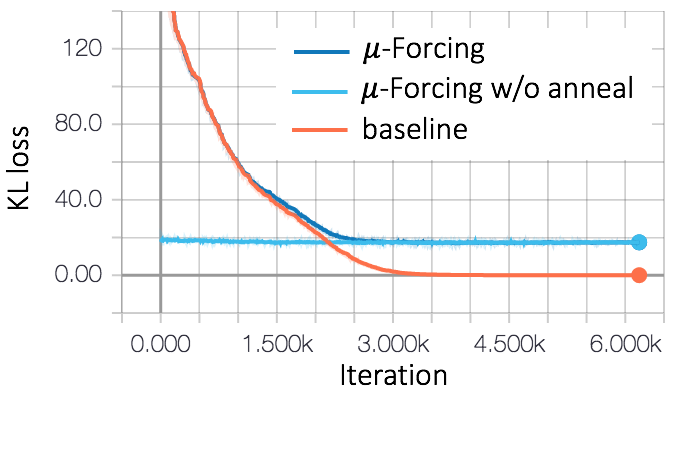}
  \end{minipage}
  }
  \subfigure[Reconstruction loss without KL annealing on COSR]{
  \begin{minipage}{4.5cm}
  \centering
    \includegraphics[width=4.5cm]{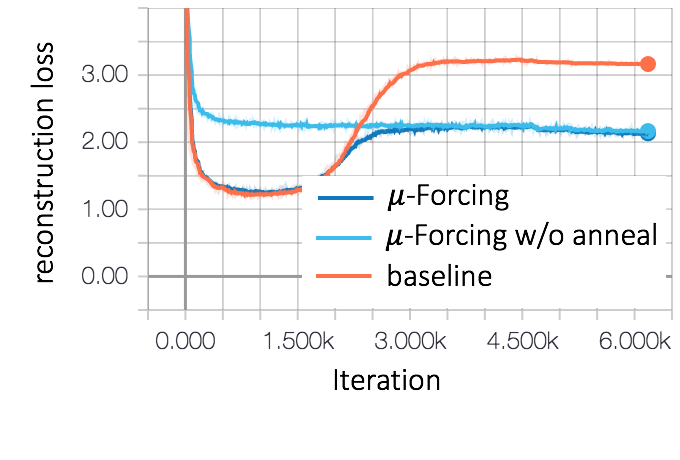}
  \end{minipage}
  }
  \caption{The KL loss and reconstruction loss of $\mu$-Forcing trained with and without KL annealing strategy on COSR [best viewed in color].}
   \label{show2}
 \end{figure}
For the second question, we focus on the effect of hyperparameter $\beta$. Based on the setup of the first experiment, we tested the proposed method on COSR dataset with $\beta$ set to 2, 3, 5, and 10, respectively. The Figure \ref{show} presents the results. As the size of $\beta$ increases, the convergency value of KL term increases while the reconstruction error decreases accordingly. These results show that the $\beta$ can flexibly control the balance between KL term and reconstruction error, and maintain tension between them. There is an important open question: what is a ``reasonable'' value of KL term \cite{hoffman2016elbo}? Ideally it should be small but non-zero. Although we didn't directly answer this question, the proposed method can control the KL term to the desired value by changing the $\beta$, which provides a feasible approach to explore this question. Note that when $\beta = 0$, the model degenerates into the baseline. While $\beta$ is too large (such as 100), it destroys the model structure and results in high KL loss and reconstruction error. We found that the model performs well in both reconstruction and generation with $\beta$ set to 2 or 3.

We also tested the performance of the proposed method without using the KL annealing training strategy. Based on the setup of the first experiment, we trained the model with the proposed method but without KL annealing on COSR dataset. The Figure \ref{show2} reports the results. In this setting, the proposed method can still hold the value of KL term, and its reconstruction error is lower than baseline on the test set. In addition, we can see that the convergency value of KL term and reconstruction error of using and not using KL annealing strategy are very close. We observed the same phenomenon with various $\beta$ on different datasets. This results indicate that KL annealing training strategy is not required when training with the proposed method.

\subsection{Comparison with Strong Baseline (Q3)}
To further evaluate the proposed method and answer the question \textbf{Q3} (How does the proposed method compare with existing state-of-the-art ones?), we firstly compared our method ($\mu$-Forcing) with several regularized-based baselines: (1) KL annealing (VRAE + KLA) \cite{bowman2015generating}. (2) VRAE with free bits (VRAE+FB). This method reserves some space of KL divergence for every dimension of latent variables \cite{kingma2016improved}. We set the reserved space for every dimension as 0.0125 in free bits (FB). (3) VRAE+FB-all. This method is similar to VRAE+FB, which reserves space for the total KL divergence instead of for for every dimension \cite{yang2017improved}. We try reserving 0.2 bits for the whole dimension space. (4) VRAE with bag-of-words loss (VRAE + BOW) \cite{zhao2017learning}. (5) VRAE with bag-of-words loss and KL annealing (VRAE + BOW + KLA) \cite{bowman2015generating,zhao2017learning}. In order to show the $\beta$ of the proposed method can flexibly control the balance between KL term and reconstruction error, we also compared the proposed method with various $\beta$ (varied from 1.5 to 3.0). In addition, we also compared the vanilla VRAE. We conducted the same word-level language modeling task on APRC dataset using VRAE. All methods use the same VRAE architecture, and the VRAE + BOW (+KLA) incorporates an additional MLP to predict the BOW\footnote{The implementation of VRAE + BOW is based on the author's source code \url{https://github.com/snakeztc/NeuralDialog-CVAE}}. For KLA, we initialize the weight with 0 and gradually increase to 1 in the first 10000 training steps. 

As with \cite{bowman2015generating,zhao2017learning}, we report the reconstruction loss (the lower is better) and KL loss on the test set. To further evaluate the generation quality, as with the evaluation in \cite{yu2017seqgan,ZhuLZGZWY18}, we use BLEU score (the higher is better) to measure the similarity degree between the generated sentences and the real sentences. We randomly sample 3,000 sentences from the test set as the references. Moreover, since the model collapse issue is a typical and classical problem for the VRAE model, we use Self-BLEU (SLEU) score \cite{ZhuLZGZWY18} (the lower is better) to evaluate the diversity of the generated sentences (other sentences generated by the model itself are taken as references). We let each model generates 3000 sentences by sampling $z$ from $\mathcal{N}(0, I)$ and calculate the average BLEU and SLEU scores (4 gram and 5 gram). The results are shown in Table \ref{exp:baseline}. 
\begin{table*}[t]
\caption{The evaluation results on APRC test set. \label{exp:baseline} }
\begin{center}
\begin{tabular}{llllllllll}
 \hline
Method  & Rec & KL & BLEU-4 & BLEU-5 & SLEU-4 & SLEU-5\\
 \hline 
VRAE & 96.48 & 0.01 & 98.16 & 94.85 & 100.0 & 100.0\\
VRAE + KLA & 96.18 & 2.57 & 90.56 & 82.99 & 94.35 & 92.18\\
VRAE + FB & 93.65 & 3.10 & 93.40 & 89.38 & 94.98 & 92.97\\
VRAE + FB-all & 91.03 & 3.88 & 89.19 & 80.55 & 94.12 & 92.01\\
VRAE + BOW & 89.32 & 6.01 & 89.28 & 80.75 & 90.39 & 86.70\\
VRAE + BOW + KLA & 86.46 & 13.28 & 83.47 & 71.77 & 81.99 & 74.95\\
\hline
$\mu$-Forcing ($\beta=1.5$) & 86.85 & 12.73 & 90.93 & 82.22 & 90.34 & 86.02\\
$\mu$-Forcing ($\beta=2.0$) & 85.44 & 16.76 & 89.09 & 78.70 & 89.01 & 83.80\\
$\mu$-Forcing ($\beta=2.5$) & 84.42 & 20.55 & 87.50 & 76.63 & 87.33 & 81.70\\
$\mu$-Forcing ($\beta=3.0$) & \textbf{83.29} & 25.15 & 84.33 & 71.99 & 85.01 & 78.26\\
 \hline
 \end{tabular}
\end{center}
\end{table*}

From the first two lines of the Table \ref{exp:baseline}, we can see that VRAE suffers from the problem of KL vanishing and serious model collapse issue. Its KL loss is approximate to zero, which means this model has collapsed into a plain language model that totally ignores the latent variable $z$. So it has high reconstruction error. Although its BLEU score is high, its SLEU scores is 100.0, it can only generate repeated sentences which we found are almost copied from the training set. With carefully adjusting the parameter values of KL annealing and free bits, the model VRAE + KLA, VRAE + FB, and VRAE + FB-all have achieved non-zero but still small KL cost. The reconstruction error of them are also very high, which indicates that these models still ignore the latent variable $z$ in most cases. Moreover, the SLEU scores of them are also much higher than other methods.

As for VARE + BOW and the $\mu$-Forcing method, all of their models solve the problem of uninformative latent variables and achieve low reconstruction error and non-zero KL cost. In addition, we can see that the KLA trick helps the VRAE + BOW achieve lower reconstruction error and higher KL cost which are in line with the results in \cite{zhao2017learning}. Its SLEU scores are lower than those of VRAE + BOW. However, its BLEU scores are lower than VRAE + BOW's. 

For the proposed method, it can be observed that as the size of $\beta$ increases, the KL cost increase while the reconstruction error decrease accordingly. These results demonstrate that the $\beta$ can flexibly control the balance between KL term and reconstruction error. In addition, although the number of parameters in $\mu$-Forcing models are only about 4/5 of the size of the VRAE + BOW based models (the VRAE + BOW incorporates an MLP to predict Bag-of-word loss), all of the $\mu$-Forcing models achieve competitive results compared with VRAE + BOW based models. From the results of generation, it can be seen that the $\mu$-Forcing ($\beta=1.5$) outperforms the VRAE + BOW model (the BLEU-4 score of $\mu$-Forcing ($\beta=1.5$) are higher than that of VRAE + BOW, at the same time, the SLEU scores and reconstruction error are lower than those of VRAE + BOW). Furthermore, although the SLEU scores of $\mu$-Forcing ($\beta=3.0$) are slightly higher than those of VRAE + BOW + KLA, its BLEU scores are higher than those of VRAE + BOW + KLA, and its reconstruction error is also lower than that of VRAE + BOW + KLA. These results show that the proposed method can solve the uninformative latent variables issue and generate high-quality and diverse sentences, which performs significantly better than VRAE + KLA, VRAE + FB and VRAE + FB-all. Furthermore, without incorporates any additional component, the $\mu$-Forcing still achieves very competitive results compared with VRAE + BOW based models.

Secondly, in order to test whether the proposed method can also be applied to other model-based methods to improve performance, we added the proposed $\mu$-Forcing regularization term to HybridVAE \cite{semeniuta2017hybrid} to compare it (Hybrid + Ours) and the auxiliary reconstruction term used in \cite{semeniuta2017hybrid} (Hybrid + Aux). We conducted the same char-level language modeling task on APRC dataset using HybridVAE. To be fair, all settings are based on the same vanilla hybrid model (Hybrid).\footnote{The implementation of HybridVAE is based on the author's source code \url{https://github.com/stas-semeniuta/textvae}} 

We report the reconstruction loss, KL loss on the test set. Similarly, the BLEU and SLEU are also shown in in Table \ref{hybrid}. From the results, we can see that the vanilla hybrid model suffers from the uninformative latent variables issue, result in high reconstruction error and KL vanishing on the test set. However, both auxiliary reconstruction term and the proposed $\mu$-Forcing term $\mathcal{L}_\mu$ can solve this issue. Furthermore, when applied our method to the vanilla hybrid model, it achieves lower reconstruction loss, SLEU scores and higher BLEU scores compare with their full model (Hybrid + Aux). These results indicate that the proposed method can be also applied to HybridVAE and further improve the performance.

\begin{table*}[h]
\caption{\label{hybrid} The evaluation results of different settings on the APRC test set.}
\begin{center}
\begin{tabular}{llllllll}
 \hline
Setting & Rec & KL & BLEU-4 & BLEU-5 & SLEU-4 & SLEU-5\\
 \hline
Hybrid  & 61.97 & 0.01 & 93.12 & 86.83 & 100.0 & 100.0\\
Hybrid + Aux & 56.01 & 12.73  & 84.28 & 70.11 & 82.18 & 73.16\\ 
Hybrid + Ours & 52.22 & 14.24 & 85.98 & 74.13 & 81.50 & 72.64 \\ 
 \hline
 \end{tabular}
\end{center}
\end{table*}

\begin{figure}[t] 
  \centering
   \includegraphics[width=4.0in]{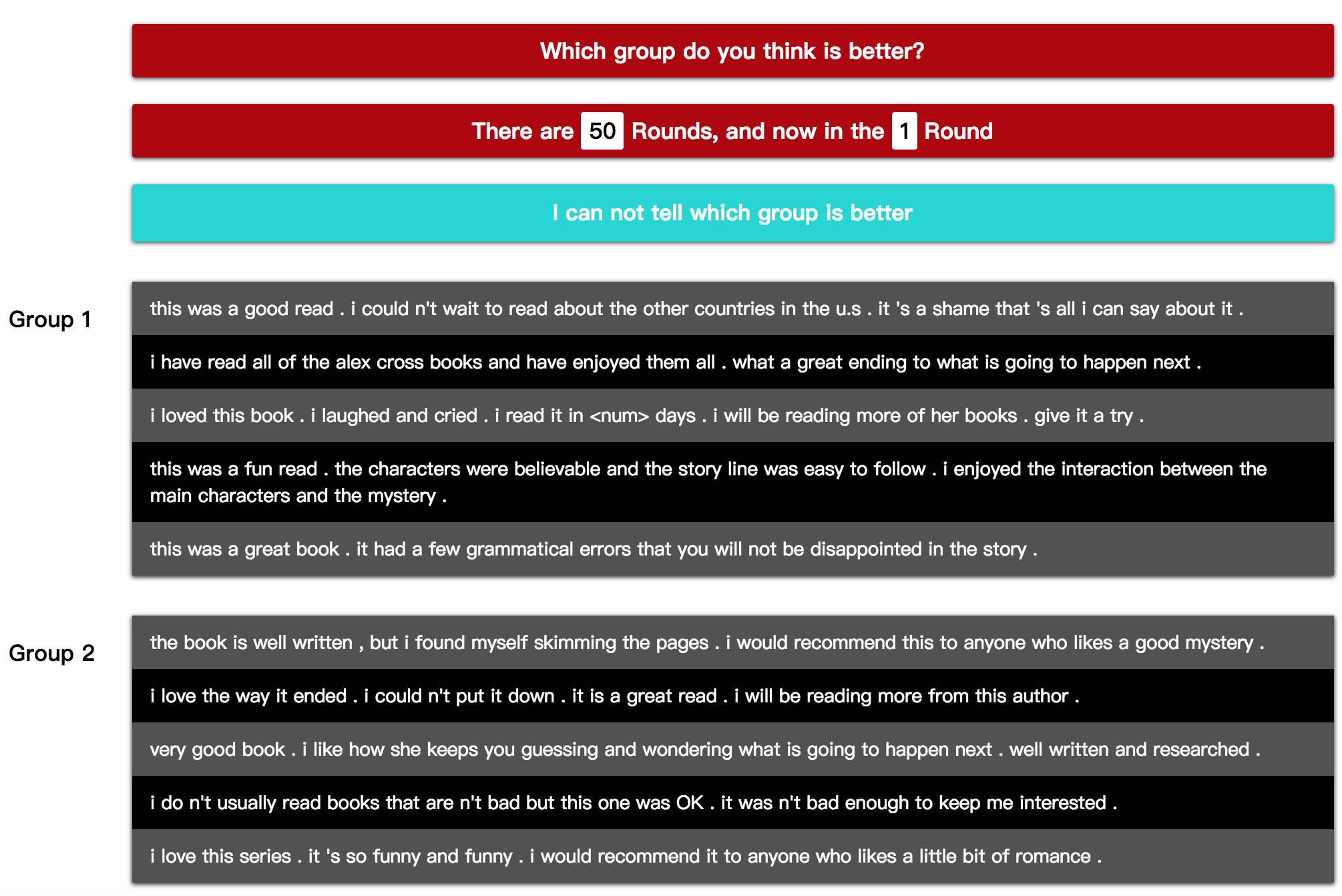}
   \caption{For each round, the human evaluation interface presents two group sentences which are generated by two different methods. Then it asks users to click on the group they think are better or choose ``I can not tell which group is better''. Each human evaluation contains 50 rounds in total.}
   \label{fig:interface}
\end{figure}

\subsection{Human Evaluation}
We conducted some human evaluations to further evaluate the proposed method on APRC task. We compared the $\mu$-Forcing with VRAE + BOW + KLA and VRAE + FB-all. All comparisons are blind paired comparisons. Because it is difficult to measure the diversity of models by comparing individual sentences directly, we compared two groups of sentences generated by different models in each round. We let each model to generate 250 sentences by sampling $z$ from $\mathcal{N}(0, I)$. For each model, the generated sentences generated were randomly divided into 50 groups, each containing 5 sentences. Then we launch a crowd-sourcing online study asking evaluators to decide which group of sentences is better (more likely to be written by human beings).

We build a user-friendly web-based environment based on Flask\footnote{http://flask.pocoo.org/} for human evaluation. The interface for human interaction is illustrated in Figure \ref{fig:interface}. For each round, the human evaluation interface presents two group of sentences which are generated by two different methods, then asks evaluators to choose the better one. Ties are permitted. A total of 15 evaluators participate in the evaluation.\footnote{All evaluators are well educated and have Bachelor or higher degree. They are independent of the authors' research group.} The results are show in Table \ref{exp:human-aprc}, and we can see that the proposed model performs best.

\begin{table}[h]
\caption{\label{exp:human-aprc} Human evaluations on APRC.}
\begin{center}
\begin{tabular}{cccc}
 \hline
$\mu$-Forcing won & VRAE + BOW + KLA won & Tied\\
 \hline 
\textbf{59\%} & 16\% & 25\% \\  
\hline\hline
$\mu$-Forcing won &  VRAE + FB-all won & Tied\\
 \hline \hline
\textbf{69\%} & 12\% & 19\% \\ 
\hline
 \end{tabular}
\end{center}

\end{table}

\subsection{Interpretable Latent Variables (Q4)}
\begin{figure*}[h]
   \centering
   \includegraphics[width=5.0in]{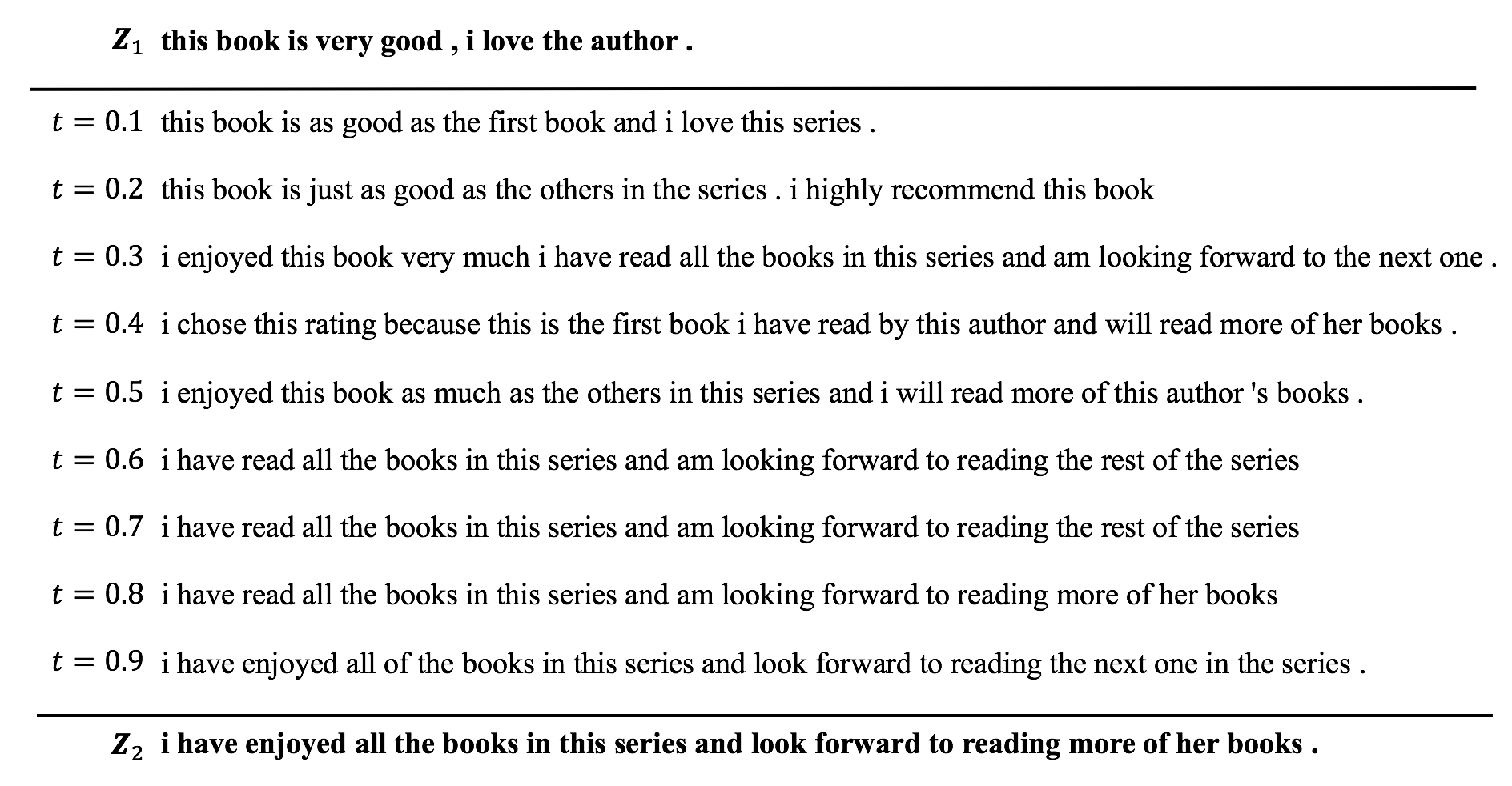}
   \caption{The results of linear interpolation in the latent space on APRC dataset. We obtained latent variables $z_1$ of the sentence in the first line and $z_2$ of the sentence in the last line. The rest of the lines report the decoding results of linear interpolation. $t$ is the interpolation parameter. In general, latent variables seem to capture the characteristics of the sentences.}
   \label{fig:figure1}
\end{figure*}
\begin{figure*}[h]
   \centering
   \includegraphics[width=3.0in]{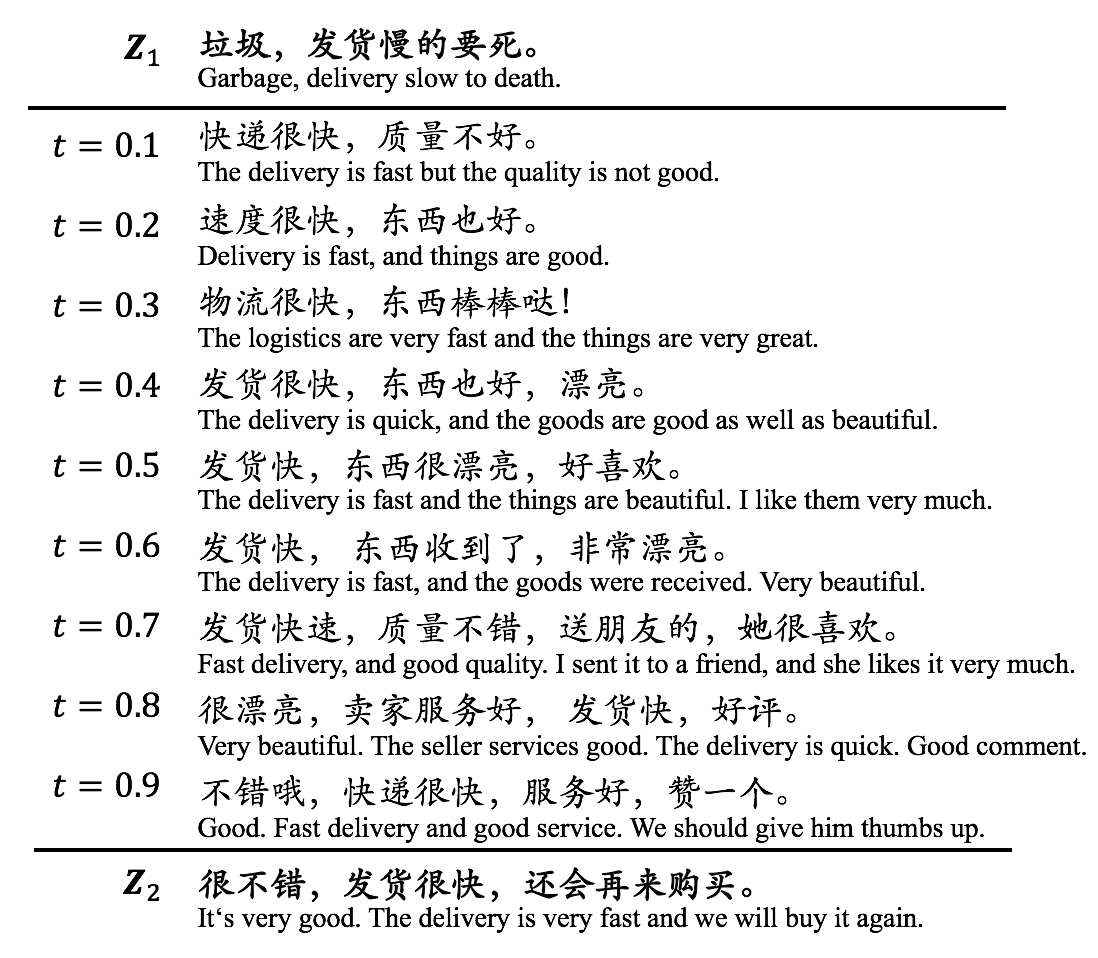}
   \caption{The results of linear interpolation in the latent space on COSR dataset. We obtained latent variables $z_1$ of the sentence in the first line and $z_2$ of the sentence in the last line. The rest of the lines report the decoding results of linear interpolation. $t$ is the interpolation parameter. In general, latent variables seem to capture the characteristics of the sentences.}
   \label{fig:afigure3}
\end{figure*}
In this experiments, we aim to answer the question \textbf{Q4} (In addition to sentence generation, the VRAE is mainly used to learn smooth and interpretable latent variables for sentence representation. Can the proposed method help the model learn interpretable latent variables?). We conducted two experiments to investigate interpretation of the latent variables.

\noindent \textbf{Homotopy}. We verified that the latent variables capture characteristics of language by \textit{homotopy} (linear interpolation) \cite{bowman2015generating} in the latent space. Given two sentences, we fed them to the encoder of VRAE and obtained their latent variables $z_1$ and $z_2$. A homotopy between $z_1$ and $z_2$ is the set of points $z_t$ on the line between them:
\begin{equation}
  z_t = z_1 \cdot t  + z_2 \cdot (1-t). \qquad  t \in [0,1]  \notag
\end{equation}
At each step $t$ of the interpolation, we generate a sentence through the decoder by feeding the latent variables $z_t$. The results on ARPC are presented in Figure \ref{fig:figure1}. We show these results on COSR in Figure \ref{fig:afigure3}. We can see that the proposed method learned dense and interpretable latent representation.

\begin{figure*}[h]
\centering
\includegraphics[width=5.0in]{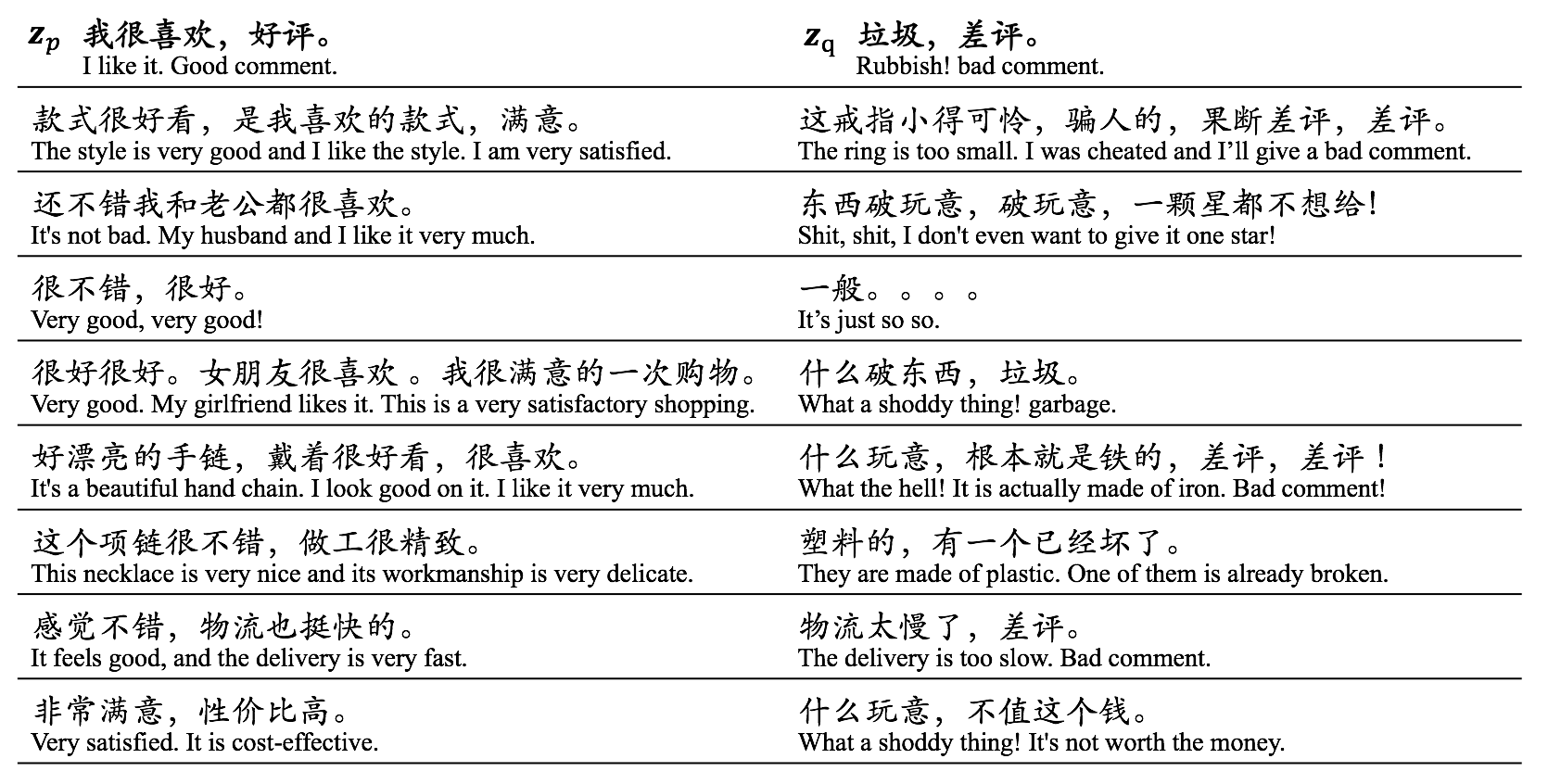}
\caption{The results of sentence sentiment transformation on COSR dataset. The first line presents two sentences, and their latent variables are $z_p$ and $z_q$ respectively. The original sentences are shown in the left columns while the sentences that are obtained after the transformation of emotions are shown in the columns on the right.}
\label{fig:afigure4}
\end{figure*}

\begin{figure*}[h]
\centering
\includegraphics[width=5.0in]{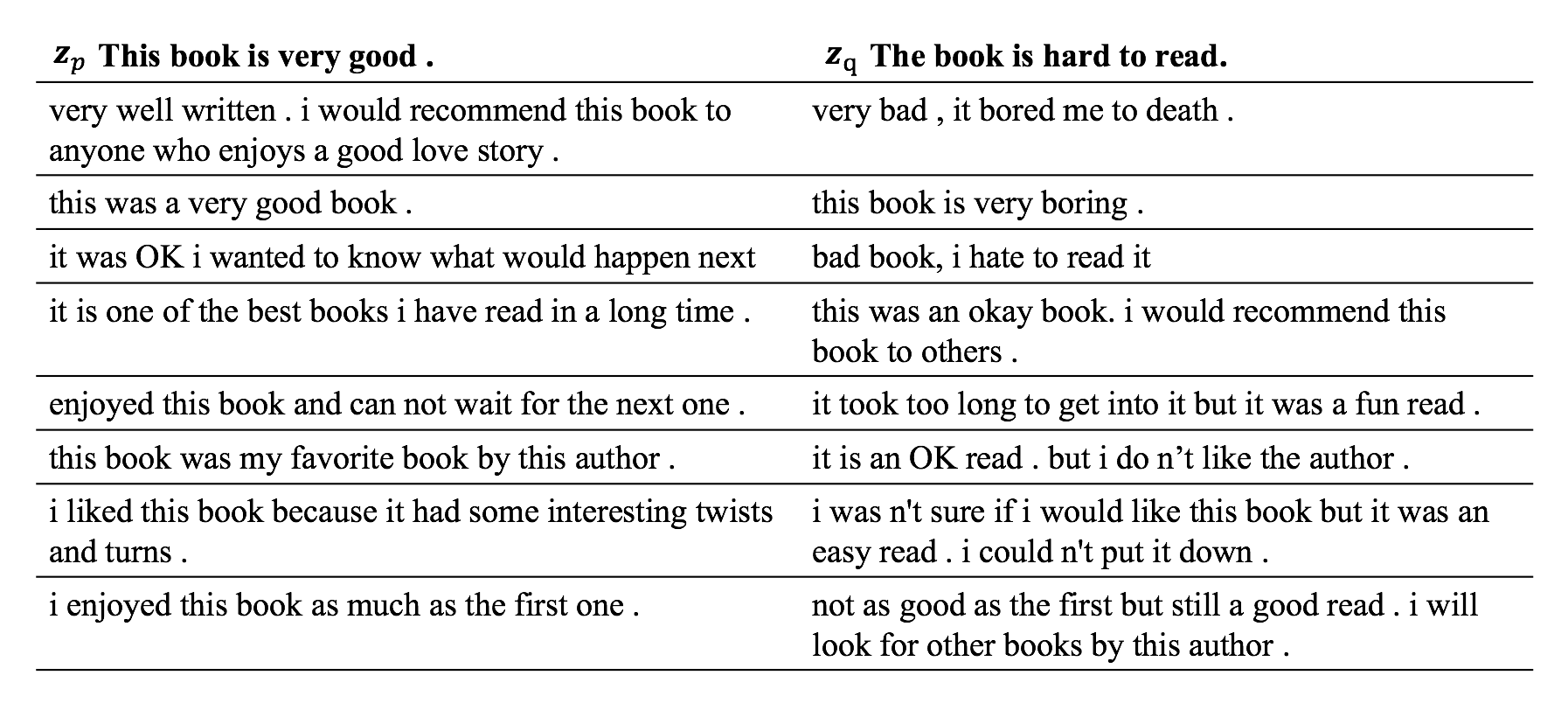}
\caption{The results of sentence sentiment transformation on APRC dataset. The first line presents two sentences, and their latent variables are $z_p$ and $z_q$ respectively. The original sentences are shown in the left columns while the sentences that are obtained after the transformation of emotions are shown in the columns on the right.}
\label{fig:figure2}
\end{figure*}

\noindent \textbf{Sentiment Transformation}. Word embedding \cite{mikolov2013linguistic} can represents words at concept level. Some language patterns are explicitly represented as linear transformations in the embedding space such as: $king - man + woman \approx queen.$ Similarly to this way, we applied linear transformation to the latent space of the proposed model for sentiment transformation. Given two sentences which are similar in content, but one emotion is positive and the other is negative. The latent variable of positive one is denoted as $z_p$ while negative one is $z_q$. Since the only difference between these two sentences is emotion, we regard the vector $z_q - z_p$ as a negative emotional vector. Let $z_a$ be a latent variable of another sentence, we turn the emotion of the sentence into a negative one by decoding the latent variable $z_b$ which is calculated as follows:
\begin{equation}
z_b = z_a + z_q - z_p.  \notag
\end{equation}
The results on APRC are presented in Figure \ref{fig:figure2}. We also show these results on COSR in Figure \ref{fig:afigure4}. We can see that most original sentences were translated into the sentences with similar content but opposite emotions. This results also demonstrate that the model learns interpretable latent variables.
\section{Conclusions}
In this paper, we further explore the reason for the uninformative latent variables of VRAE. To address this issue, we propose an effective regularizer based approach. The proposed method introduces a mild constraint on the $\mu$ of $q(z | x)$ to force the model to find a non-trivial solution where the learned latent variables $z$ contain useful information, which can perform well without using other strategies, such as KL annealing. The experiments show that the proposed method outperforms several strong baselines and can flexibly and stably control the trade-off between the KL term and the reconstruction term of VRAE during training, making the model learn interpretable latent variables and generate diverse meaningful sentences.

In future work, we plan to utilize the learned latent representations to improve the semi-supervised learning on NLP tasks, such as text classification and sentiment detection. Also, it would be interesting to apply the proposed method to other VAE based models for image generation.

\begin{acks}
This work is supported by  the National Key R\&D Program of China under contract No. 2017YFB1002201, the National Natural Science Fund for Distinguished Young Scholar (Grant No. 61625204), and partially supported by the State Key Program of National Science Foundation of China (Grant Nos. 61836006 and 61432014). 
\end{acks}
\bibliographystyle{ACM-Reference-Format}
\bibliography{main}

\end{document}